\documentclass[letterpaper, 10 pt, conference]{ieeeconf}  

\IEEEoverridecommandlockouts                              

\usepackage{lipsum}
\usepackage[utf8]{inputenc}
\usepackage{amsmath,amssymb,amsfonts,bm}
\usepackage{mathrsfs}  
\usepackage{xcolor}
\usepackage{algorithmic}
\usepackage{graphicx}
\usepackage{textcomp}
\usepackage{xcolor}
\usepackage{hyperref}

\makeatletter
\newcommand*\titleheader[1]{\gdef\@titleheader{#1}}
\AtBeginDocument{%
  \let\st@red@title\@title
  \def\@title{%
    \bgroup\normalfont\large\centering\@titleheader\par\egroup
    \vskip1.5em\st@red@title}
}
\makeatother

\title{\LARGE \bf
LOL: Lidar-only Odometry and Localization in 3D point cloud maps*
}
\titleheader{2020 IEEE International Conference on Robotics and Automation - arXiv prepint version}

\author{Dávid Rozenberszki and András L. Majdik$^{1}$
\thanks{*The research reported in this paper was supported by the Hungarian Scientific Research Fund (No.\ NKFIH OTKA KH-126513) and by the project: Exploring the Mathematical Foundations of Artificial Intelligence 2018-1.2.1-NKP-00008.
}
\thanks{$^{1}$Dávid Rozenberszki and András L. Majdik are with the Machine Perception Research Laboratory,
        MTA SZTAKI, Hungarian Academy of Sciences, Institute for Computer Science and Control, 1111 Budapest, Hungary
        {\tt\small \{rozenberszki,majdik\}@sztaki.hu}}%
}

\begin{document}

\maketitle
\thispagestyle{empty}
\pagestyle{empty}

\begin{abstract}
In this paper we deal with the problem of odometry and localization for Lidar-equipped vehicles driving in urban environments, where a premade target map exists to localize against.
In our problem formulation, to correct the accumulated drift of the Lidar-only odometry we apply a place recognition method to detect geometrically similar locations between the online 3D point cloud and the a priori offline map.
In the proposed system, we integrate a state-of-the-art Lidar-only odometry algorithm with a recently proposed 3D point segment matching method by complementing their advantages. 
Also, we propose additional enhancements in order to reduce the number of false matches between the online point cloud and the target map, and to refine the position estimation error whenever a good match is detected.
We demonstrate the utility of the proposed LOL system on several Kitti datasets of different lengths and environments, where the relocalization accuracy and the precision of the vehicle's trajectory were significantly improved in every case, while still being able to maintain real-time performance.

\end{abstract}



\section*{SUPPLEMENTARY ATTACHMENTS}
The source code of the proposed algorithm is publicly available at: \url{https://github.com/RozDavid/LOL}
A video demonstration is available at: \url{https://youtu.be/ektGb5SQGRM}

\section{INTRODUCTION}

In this paper, we deal with the challenges of globally localizing a robot in urban environments using exclusively Light Detection and Ranging (Lidar) measurements. Parallel a Lidar odometry and mapping algorithm is running to calculate the estimated consecutive poses from the streaming Lidar point clouds in real-time, while the global position updates are recognized against a previously known point cloud of a target map---accepted as ground-truth--- to cancel the built-up drift. 

Our motivation stems from the certain need for Lidar based sensors in autonomous navigation and driving as these types of measurements can provide relatively high precision data, with constant irrespective of the distance measured. Also compared to simple RGB or RGB-D images it is generally less sensitive to seasonal and adversarial weather changes. In many applications though, they complete each other, as well as with other sensors (e.g. Inertial Measurement Unit (IMU), wheel odometry, radar) for the best performance in machine perception. However, for true redundancy, these sub-systems must work independently as well, to avoid dangerous situations from sensor malfunction. Using solely Lidar data has its difficulties also, as they operate on almost one scale lower frequency as cameras and even two or three lower as IMU sensors. Moreover, their scans are sparse in points, but usually still computationally expensive to process them in real-time.


In case of pure Lidar odometry the LOAM \cite{loam} algorithm scores among the highest in term of translational and rotational errors on the Kitti Vision Benchmark \cite{kitti} with 2-axis Lidars such as Velodynes.
This method estimates the 6 DoF (Degree of Freedom) displacement of a moving Lidar with very low drift on short trajectories by frame-to-frame comparison of the consecutive scans, and by fine matching and registration of the consecutive point clouds into a coordinate system aligned online map. 
However, in case of long trajectories and since the drift is continuously accumulated a significant error could build up in the position estimation, c.f. Fig. \ref{fig:loam_maps} top row.

In our problem formulation, to correct the accumulated error in the odometry we apply a place recognition method to detect geometrically similar locations between the online and the a priori offline 3D point cloud map. Our approach is opposite to Simultaneous Localization and Mapping (SLAM) systems, where the goal is to detect loop-closure situations whenever the vehicle returns to a previously visited location in the environment, within the online map. For the place recognition frontend we chose to integrate the SegMap \cite{segmap} method, that is a state of the art algorithm for the extraction and matching of 3D point cloud segments. 

Our main contribution is the integration and adaptation of the LOAM and SegMap algorithms into a novel solution, creating thus a Lidar-only Odometry and Localization (LOL) method that eliminates the need of any other supplementary sensor, e.g., IMU, wheel encoder, and satellite-based Global Positioning System (GPS). Furthermore, we included some additional improvements in the acceptance of correct matches, by applying further geometrical constraints complementing the feature similarity ones. Namely, we apply a RANSAC based geometric verification, and once a good match is detected between the online measurements end the target map, we only search for similar 3D Lidar segments (with relaxed similarity constraints) in the neighbourhood of the current location defined by the location uncertainty. Also, we only use the shift between the target map and the online source segments centroids as a prior, and we refine the final transformation by applying a fine-grained ICP matching between the two point clouds. We tested the proposed algorithm on several Kitti datasets, c.f. Fig. \ref{fig:loam_maps} bottom row, and found a considerable improvement in term of precision without a significant computational cost increase. 

To summarize, this paper advances the state-of-the-art with the following contributions:
\begin{itemize}
    \item We present a novel Lidar-only odometer and Localization system by integrating and complementing the advantages of two state of the algorithms. 
    \item We propose a set of enhancements: (i) a RANSAC-based geometrical verification to reduce the number of false  
    matches between the online point cloud and the offline map; and (ii) a fine-grained ICP alignment to refine the relocalization accuracy whenever a good match is detected.
    \item We publicly release the source code of the proposed system.


\end{itemize}{}

The remainder of the paper is organized as follows: related work is presented in Section \ref{related}, the details of the proposed system are described in Section \ref{proposed_system}, followed by experiments in Section \ref{experiments} and conclusions in Section \ref{conclusion}.

\section{RELATED WORK} \label{related}

\begin{figure*}[!h]
\centerline{\includegraphics[width=0.72\textwidth]{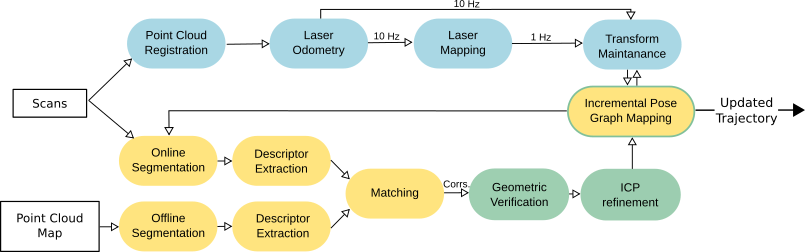}}\par
\caption{Flowchart of the complete Lidar Odometry and Localization algorithm, where the blue color represents the parts originated from the LOAM, orange color from the SegMap algorithm and the green modules are our contribution. When only the border is green as in the Incremental Pose Graph Mapping Module we modified the original SegMap method to incorporate the relocalization in the global map.}
\label{fig:loam-segmap_flowchart}
\end{figure*}

Lidar sensors are robust and precise ranging sensors for a wide range of weather conditions and distances, thus making it crucial for autonomous navigation problems. 
The related scientific literature can be grouped along with the two sub-problems addressed in this paper. Namely, the odometry measurements are calculated with the frequency of the scans providing a continuous pose estimate, while the re-localization happens occasionally against a previously known target ground truth map, only when an adequate number of similarities are detected.

As the ICP (Iterative Closest Point) algorithms become the dominant method for the alignment of three-dimensional models based on geometry \cite{essential_icp}, it is used as the initial step of most Lidar based odometry calculation problems. While typical ICPs assume that the points in the cloud are measured simultaneously and not sequentially (practically it is rather 4D than 3D) as most available rangefinders operate, it is difficult to avoid accumulated tracking error for continuous motion. For fast-moving objects such as autonomous vehicles an additional second step is incorporated into the algorithm to remove the distortion with know velocity measured from IMU sensors in \cite{imu_icp}, with multiple sensors using extended Kalman Filter (EKF) \cite{ekf_icp} or calculating the velocity directly from the Lidar scans without the aid of additional sensors such as in \cite{vicp}. Additionally, the precision of the estimate can be further improved by incorporating a lower frequency mapping and geometric point cluster matching method into the odometry \cite{zlot}, calculating loop closures in \cite{bosse} or representing the surface with Implicit Moving Least Squares (IMLS) \cite{imls_icp}. 

There is an even bigger difference in approaches to the problem of localization. On the one hand, one can explore curb detection based methods. Wang et al. \cite{curb} approached the solution by detecting the curbs at every single frame, densified with the projection of the previous detections into the current frame, based on the vehicle dynamics, applying the beam model to the contours and finally matching the extracted contour to the high-precision target map. Another approach with similar idea by Hata et al. \cite{curb_mc} is simply using these curbs complemented with a robust regression method named Least Trimmed Squares (LTS) as the input of the Monte Carlo Localization algorithm \cite{monte-carlo}. These methods share the same strengths in the independence of scale, as they both operate in the given area of the target map around the position estimate.

Numerous research works are tackling the problem of localization based on the comparison of local keypoint features of the local- and target map. Bosse and Zlot \cite{bosse-zlot} are using 3D \textit{Gestalt} descriptors for the description of point cloud directly and vote their nearest neighbours in the feature space with a vote-matrix representation. Range-image conversion approaches are also present with features such as Speeded Up Robust Features (SURFs) in \cite{surf1, surf2}, but the closest to our segment matching approach are the ones including features such as Fast Point Feature Histogram (FPFH) \cite{fpfh}.

A more general approach of the problem when global descriptors are used for the local point clouds and compared to the database of target map segments for instance in \cite{rohling} or at Magnusson et al. \cite{magnusson}.
A higher-level representation of the point cloud segment of object-level was presented by Finman et al. \cite{object}, but only for RGB-D cameras and a small number of objects. Most global feature-based descriptors though operate on a preprocessed, segmented point clouds of the local and global frame, describing the objects individually, and requiring a good prior segmentation algorithm. On well-defined segments, there are existing algorithms that are not relying on simplistic geometric features such as Symmetric Shape Distance in \cite{ssdistance}. The fastest emerging trend of deep learning is also applicable in point cloud environments such as segments and can be used for description or reconstruction as well. The learning-based method opens a completely new window to solving the above problems in a data-driven fashion. Weixin et al. in \cite{l3net} presented L3-Net a 3D convolution RNN directly operating on the point clouds and tested to exceed the results of the geometry-based pipelines on various environments. Zeng et al. proposed the 3DMatch in \cite{3dmatch}, one of the first data-driven CNN for localization in the 3D segment space. Autoencoder networks are presented in \cite{autoe1, autoe2} for localization and reconstruction as well. A different and interesting approach is the DeepICP \cite{deepicp}, an end-to-end 3D point cloud registration framework replacing standard RANSAC based ICP algorithms for matching the local and target clouds, resulting in a comparably better performance than previous state-of-the-art geometry-based methods. 

The introduced related works were listed without being exhaustive. 
These works used very different approaches reaching high performance, but only focusing on a section of our problem. In most real-world scenarios in autonomous driving applications, target maps are or will be available to localize against, and should be utilized if one can find true matches with high confidence. Though the necessity is given, the field lacks in comprehensive Lidar based odometry and localization methods for real-world applications.

\section{PROPOSED SYSTEM} \label{proposed_system}
In this section, we present the building blocks of the proposed method for our Lidar-only odometry and Localization (LOL) solution. First, we introduce the general structure and the data flow between the different components and their connections to help the reader understand the logic behind our approach.
Next, we briefly present the Lidar Odometry and Mapping (LOAM) \cite{loam}, and the Segment Matching (SegMap) \cite{segmap} algorithms.
Finally, we highlight the changes we've made to integrate them, and we present further enhancements to calculate the position updates from the matching segments.

\subsection{System architecture} \label{system_architecture}
We visualized the data flow and the logical diagram of our architecture in Fig. \ref{fig:loam-segmap_flowchart}. Here the modules shown by the blue color are originated from the LOAM algorithm, the orange ones from the SegMap algorithm and finally the modules represented with green color are our contribution and additions to integrate these into the proposed solution. 

As Fig. \ref{fig:loam-segmap_flowchart} indicates, two inputs are required for our LOL algorithm to operate with an additional assumption for start position before first localization. 

Firstly, we rely on the target map in the form of a 3D point cloud, where we would like to localize our vehicle. For starting the algorithm we first take the point cloud map, segment it and describe the created segments in a way that will be detailed later in Subsection \ref{localization}. This way we create a database of segments where we can search by the global position of the segment centroids and access their feature vector values to compare against our local map segments for calculating the correspondence candidates and later matches. 

Secondly, when the target map is set, we can start streaming the Lidar scans for the rest of the sequence. The scans are forwarded to the Point Cloud Registration module, the first part of the LOAM algorithm, where it is filtered, registered and transmitted to the following modules. Here the odometry is calculated with maintaining the pose transform of the vehicle in the global reference frame by a prior odometry calculation, then a refinement of the pose with the mapping module all detailed in Subsection \ref{loam} and shown on Fig. \ref{fig:loam-segmap_flowchart}. The very same scans are used for building the local map of the vehicle but only used for this purpose when there is an odometry estimate available for the local frame in the global space. These scans are densified by utilizing the last \textit{k}---a predefined parameter---scans with their belonging odometry estimate, then segmented and described the same way as previously the target map, only in an online manner. 

The last orange module of Fig. \ref{fig:loam-segmap_flowchart} the segmented local map compared to the local area of the target map and the matches are found between the most similar segments. The similarity is defined in the absolute distance of n dimension feature vectors, where \textit{n} depends on the descriptor module of SegMatch \cite{segmatch} or SegMap. The correspondence candidates are here filtered by our two-stage RANSAC based method to filter the outliers, thus false positive matches are removed before calculating a general transformation matrix for the global relocalization. 

Finally, to calculate a more accurate localization after the prior join of segment centroids, we apply an ICP refinement to the local and target segment point clouds as it will be detailed in \ref{our_contribution}. This final and combined update transformation matrix is inserted into the incremental pose graph mapping module of the SegMap algorithm, simultaneously feedbacked for the creation of the local map and published as the true global trajectory of the vehicle.

\subsection{Lidar-only odometry}\label{loam}

The problem is addressed for the estimation of the ego-motion of a pre-calibrated 3D Lidar, based on the perceived streaming point clouds. The additional mapping function is not used to update the target maps, though fully taken advantage of for refining the odometry estimate, as will be detailed later in this subsection. We have to note, that this subsection will be only a brief introduction for the complete LOAM algorithm, but can be found in the original paper.

\begin{figure*}[ht!]
    \begin{center} \begin{tabular}{c@{\hspace{10mm}}c@{\hspace{10mm}}c@{\hspace{10mm}}}
        \includegraphics*[width=0.23\linewidth]{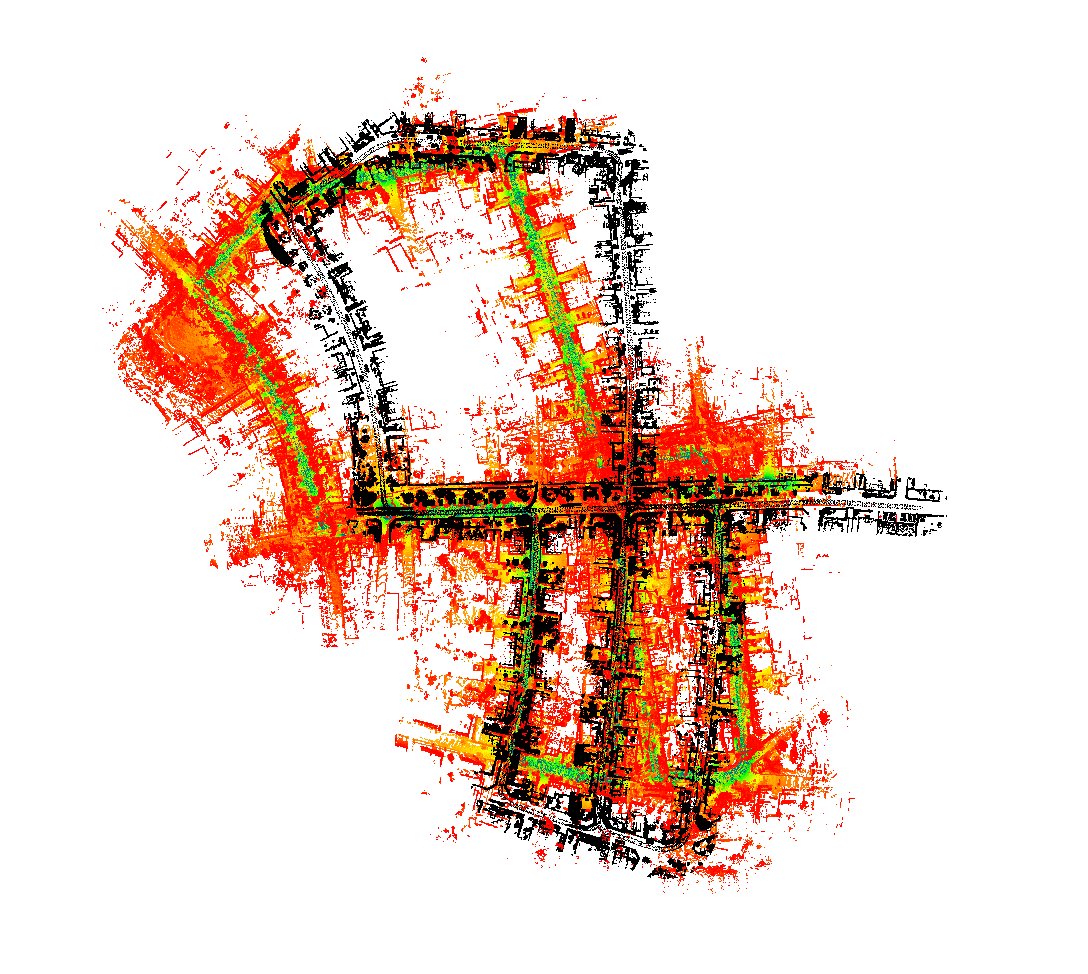} &
        \includegraphics*[width=0.23\linewidth]{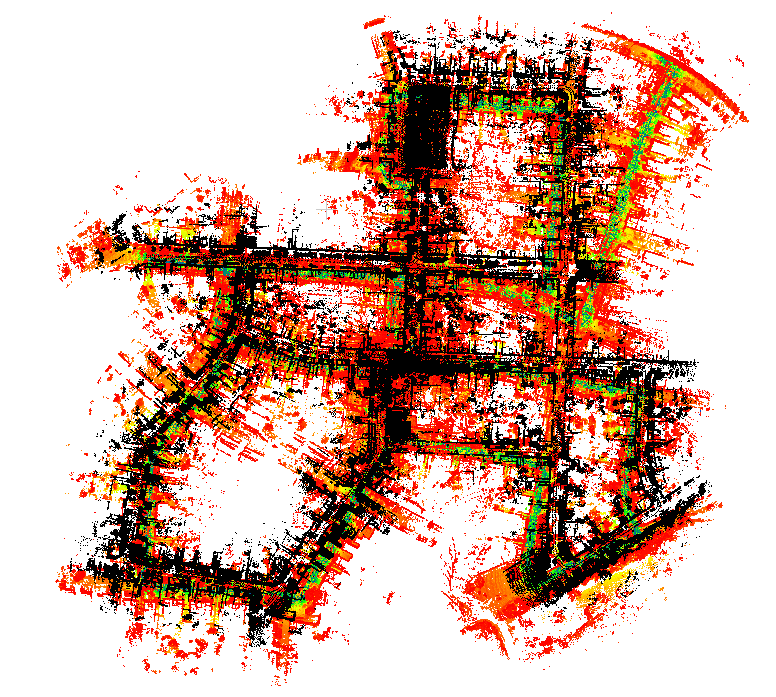} &
        \includegraphics*[width=0.23\linewidth]{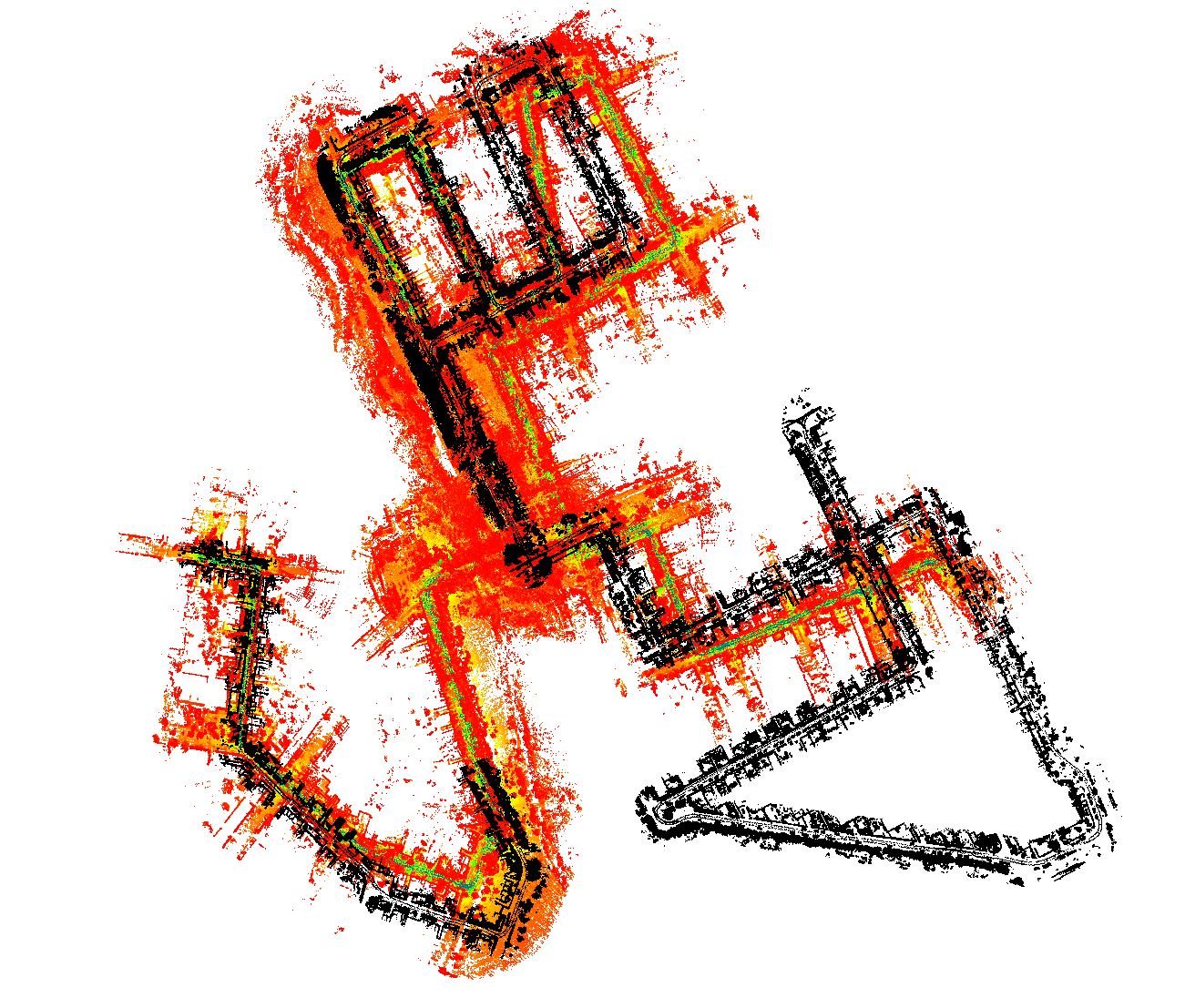} \\
        \includegraphics*[width=0.23\linewidth]{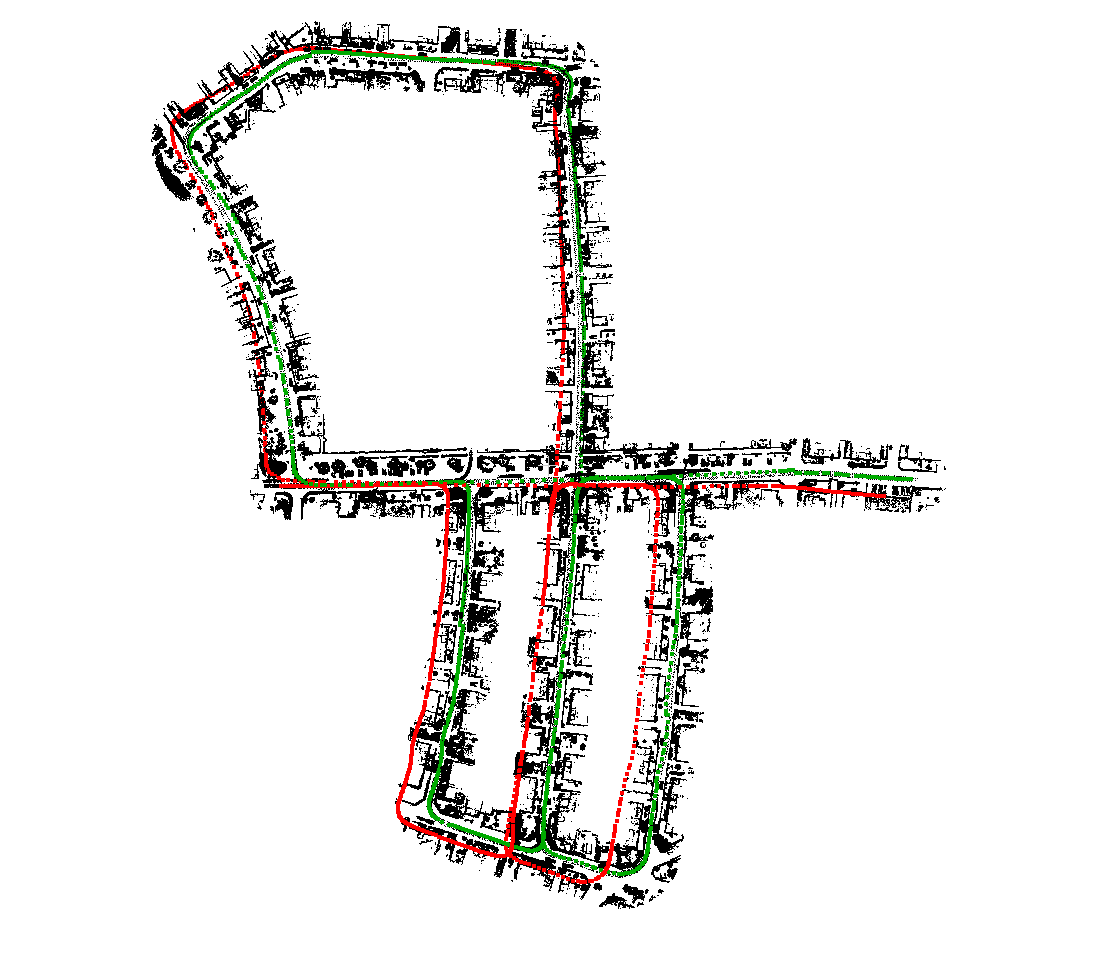} &
        \includegraphics*[width=0.23\linewidth]{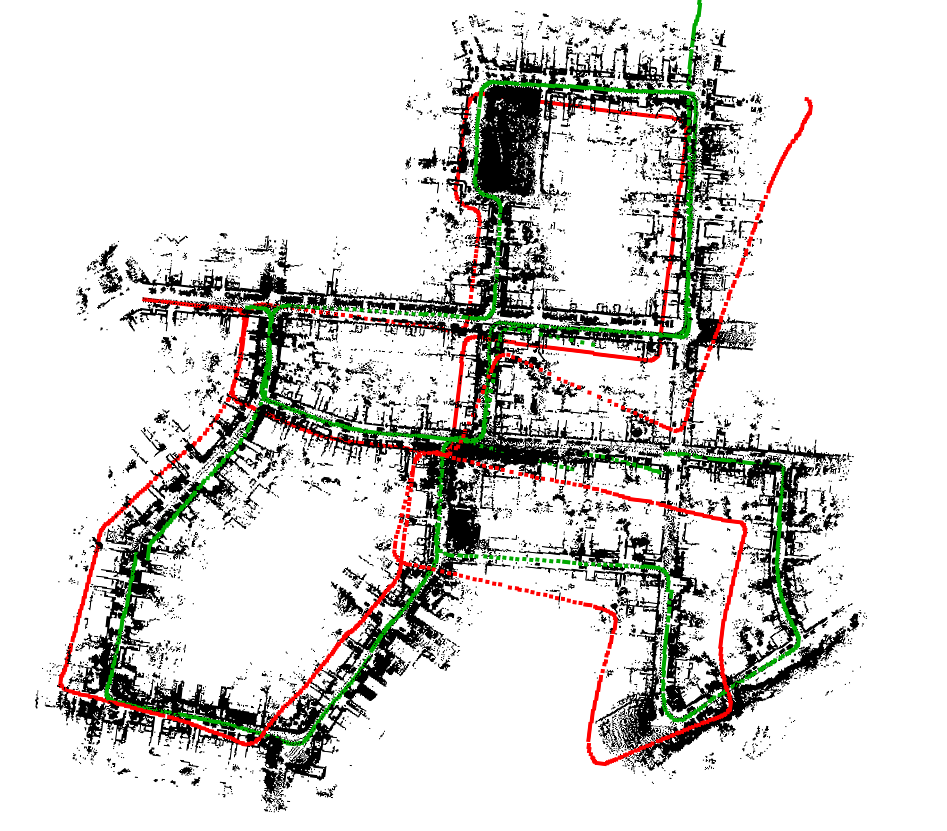} &
        \includegraphics*[width=0.23\linewidth]{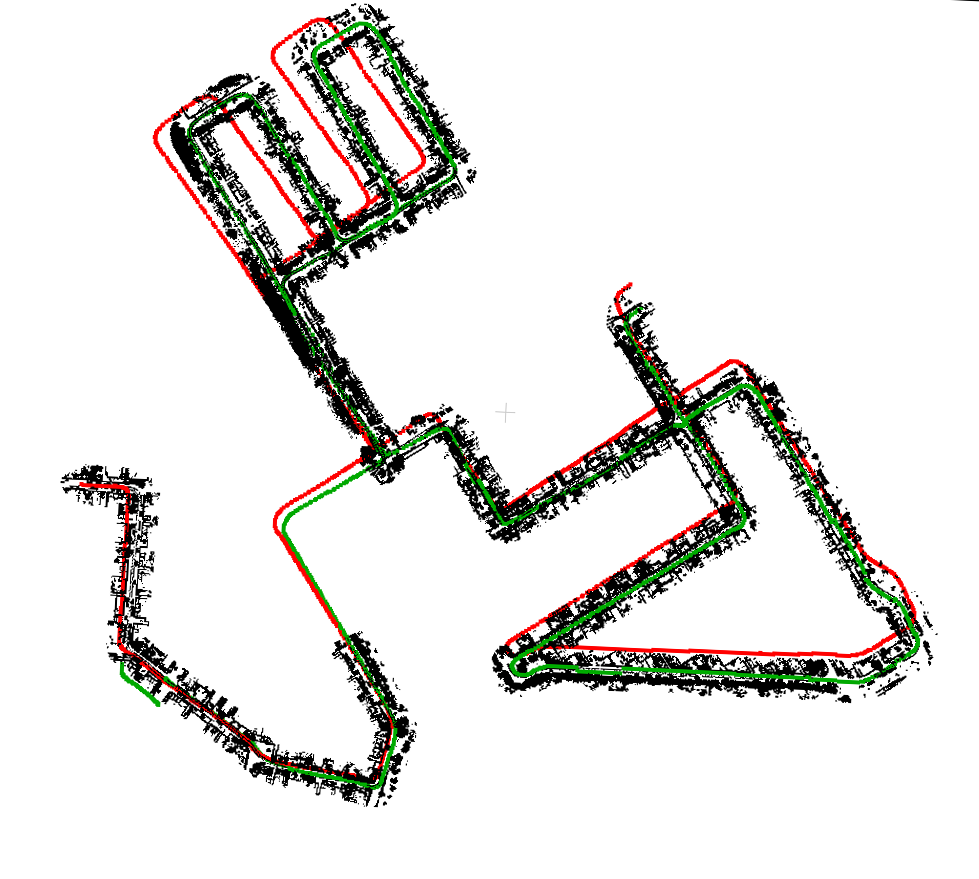} \\
         (a) & (b) & (c)\\
    \end{tabular}
        \caption{Illustration of results and comparison between the estimated trajectories tested on various length Kitti \cite{kitti} datasets: (a) Drive 18, 04:36 minutes, $\approx 2200 m$ (b) Drive 27, 07:35 minutes, $\approx 3660 m$ (c) Drive 28, 08:38 minutes, $\approx4125 m$. \textbf{Top row:} result of the LOAM algorithm, where the ground truth map is visualized with black points, while the self built map points on the green-red scale according to the vertical height. \textbf{Bottom row:} comparison between the trajectory obtained by applying the proposed LOL algorithm (green line) and the original LOAM trajectory shown (red line) with respect to the ground truth map.
        }
    \label{fig:loam_maps}
    \end{center}
\end{figure*}

As it can be seen in Fig. \ref{fig:loam-segmap_flowchart}, the first challenge is to calculate a suitable set of feature points from the received laser scans. Let $\mathcal{P}_k$ be a received point cloud scan for the sweep $k \in Z^+$. For the Lidar ego-coordinate system ${L}$ the algorithm was modified to use a more typical coordinate axes convention of $x$ forward, $y$ left and $z$ up right-hand rule. The coordinates of the point $i$, where $i \in \mathcal{P}_k$ in ${L_k}$ are denoted as $\boldsymbol{X}_{k,i}^L$. 
We would like to select the feature points that are on sharp edges and planar surfaces, by calculating the smoothness of the local surface. The selected feature points are filtered by region to avoid over crowded areas, unreliable parallel surfaces and  within the accepted threshold of $c$ smoothness values $c$.
When the accepted feature points of the reprojected $\mathcal{P}_k$ and $\mathcal{P}_{k+1}$ are available, their feature points are stored in a 3D KD-tree with their calculated $c$ values and separated as planar or edge points. The closest neighbours of feature points are then selected from the consecutive scans to calculate the edge lines of two or more points or planar patches for a minimum number of three planar feature points. Then for verification of the correspondence, the smoothness of the local surface is checked based on the same smoothness equation.

For the 6DOF pose transform LOAM solves a nonlinear optimization problem to calculate the rigid motion $\mathbf{T}^L_{k+1}$ between $[t_{k+1}, t]$.
With the converged or after the maximum iteration number terminated optimization $\mathbf{T}^L_{k+1}$ contains the Lidar ego-motion between $[t_{k+1}, t_{k+2}]$ the final step of the LOAM algorithm, the Mapping registers the scans $\mathcal{P}_k$ in \{W\} for all previous scans, thus defining $\mathcal{Q}_k$ point cloud on the map.
We use the same smoothness equation on feature point cloud $\mathcal{Q}_k$ for refining the odometry estimate by the registered map, but with a 10 times lower frequency and 10 times higher number of points than we used previously for the odometry cloud $\mathcal{P}_k$. 

Finally, to evenly distribute the points in the map the clouds are voxelized into a 5 cm grid and the odometry poses are updated by the mapping frequency and broadcasted to the localization part of our algorithm. 

\subsection{3D place recognition} \label{localization}
As seen on Fig. \ref{fig:loam_maps}, LOAM odometry and consequently its mapping performance is outstanding along short trajectories. However, in case of longer sequences, a significant drift is accumulated, that needs to be cancelled for reliable vehicle position estimates. For this purpose, we choose the SegMap algorithm with additional descriptors used from the Segmatch algorithm \cite{segmatch} to localize against a known target map. These algorithms share a common approach in the localization problem as using both the local and target maps the same way, thus will be introduced together. The only difference in the two methods are the feature values and dimensions of the segments as it will be detailed later in this subsection. 

Fig. \ref{fig:loam-segmap_flowchart} shows the complete pipeline including the modular design of the SegMap algorithm. It first extracts the pointcloud after a voxelization and filtering step then describes the segments from the 3D Lidar point clouds and matches these segments to the offline segmented and described target map. Due to the modular design of this framework, it was straightforward to implement the different methods of feature extraction and description into the system, such as incorporating the novel 3D point cloud Convolutional Neural Network (CNN) descriptor of SegMap.


Here we tested and used two different methods, the first based on the original SegMatch algorithm, then the second proposed in the SegMap paper as previously mentioned. 
On a given segment of $\mathcal{C}_k$ an optional length feature vector is created containing the different type of descriptors: $\mathscr{f}_i=[\mathscr{f}^1, \mathscr{f}^2, \dots, \mathscr{f}^m]$:

\begin{itemize}
    \item $\mathscr{f}_1$ \textit{Eigenvalue based} method calculates the eigenvalues of the segment's points and combined in a 1x7 vector. Here  the \textit{linearity}, \textit{planarity}, \textit{scattering}, \textit{omnivarence}, \textit{anisotropy}, \textit{eigenentropy} and \textit{curvature change} measures as proposed in \cite{eigen_features}. 
    \item $\mathscr{f}_2$ A CNN based approach to achieve a data driven descriptor. First the voxelized segment is resized to a fixed 32x32x16 input size, then forwarded to a simple convolutional network topology. Three 3D convolutional and max pooling layers are followed by two fully connected (FC) layers with dropout. Also the original scale of input segment is passed as an additional parameter to the first FC layer to increase robustness. The final descriptor vector of 1x64 size is obtained by the activations of the last of the two FC layers. The weights of the network are trained and optimized as detailed in the original paper \cite{segmap}. 
\end{itemize}


Finally, on the results the learning-based classifier is responsible for building the final list of match candidates based on the feature vector similarities. 

\subsection{Localization update and further enhancements}\label{our_contribution}

The previously introduced modules were thoughtfully chosen to work together and complete each other, hereby we present the details to incorporate the segment matching recognitions and the real-time odometry estimates into our solution. 
This following section is also meant to present our contribution regarding the refined relocalization performance and filtering of the false-positive matches with an additional advanced RANSAC based geometric verification. 

First, we extended the geometric verification step of the original SegMap algorithm with additional filtering for location. We search in the database of segments only within an adjustable threshold of centroid distance from our odometry position, this way reducing the putative correspondence candidates. After the matches are found in this reduced pool of segments we check the consistency of pairs with the following criteria. The absolute difference of translation vectors between the different match pair centroids can not exceed a given parameter, this way filtering the clear outliers with a minimal computing capacity. 
Next, we create a pointcloud of the remaining matching segment centroids and through a RANSAC iteration we try to align the target and source clouds. If the algorithm converged we successfully filtered the remaining outliers and now we have a high confidence in true positives. On these matches we can calculate the required 6DOF update transformation between the estimated and the true global pose of the vehicle.
The performance of our filtering method can be seen in Table \ref{tab:ransac_table}. 

\begin{figure}[ht!]
    \begin{center} \begin{tabular}{c@{\hspace{1mm}}c@{\hspace{1mm}}}
        \includegraphics*[width=0.44\linewidth]{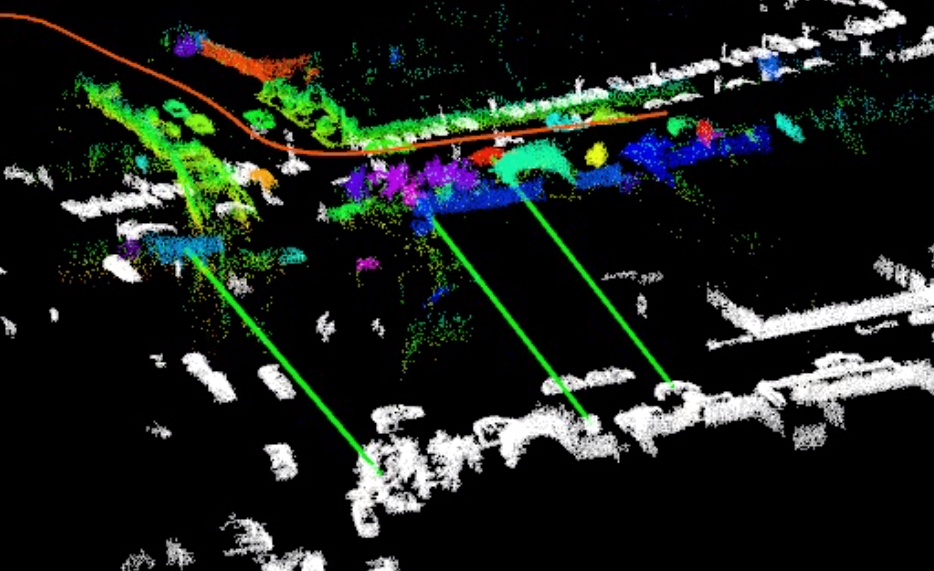} &
        \includegraphics*[width=0.44\linewidth]{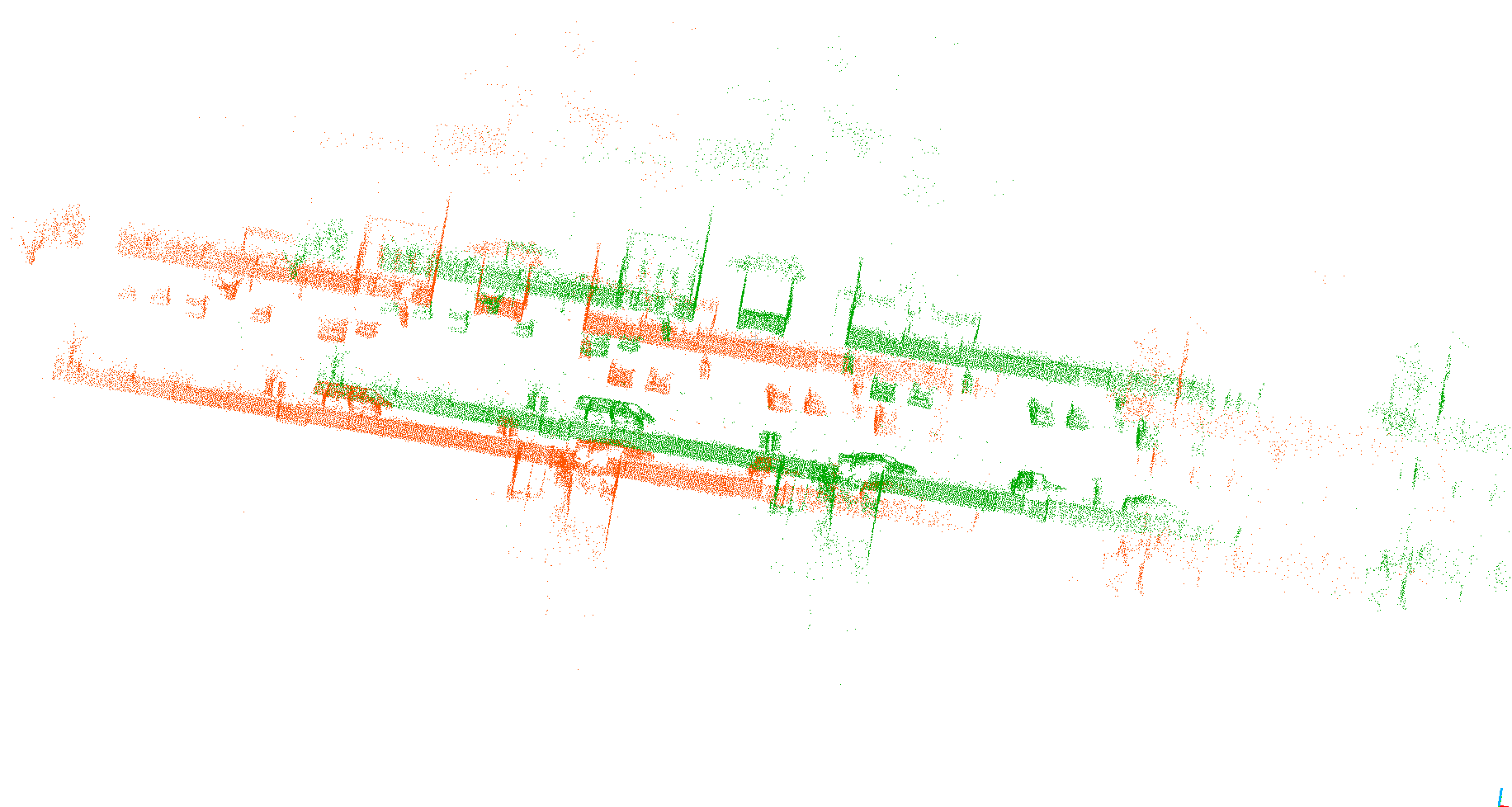} \\
        (a) & (b) \\
        \includegraphics*[width=0.44\linewidth]{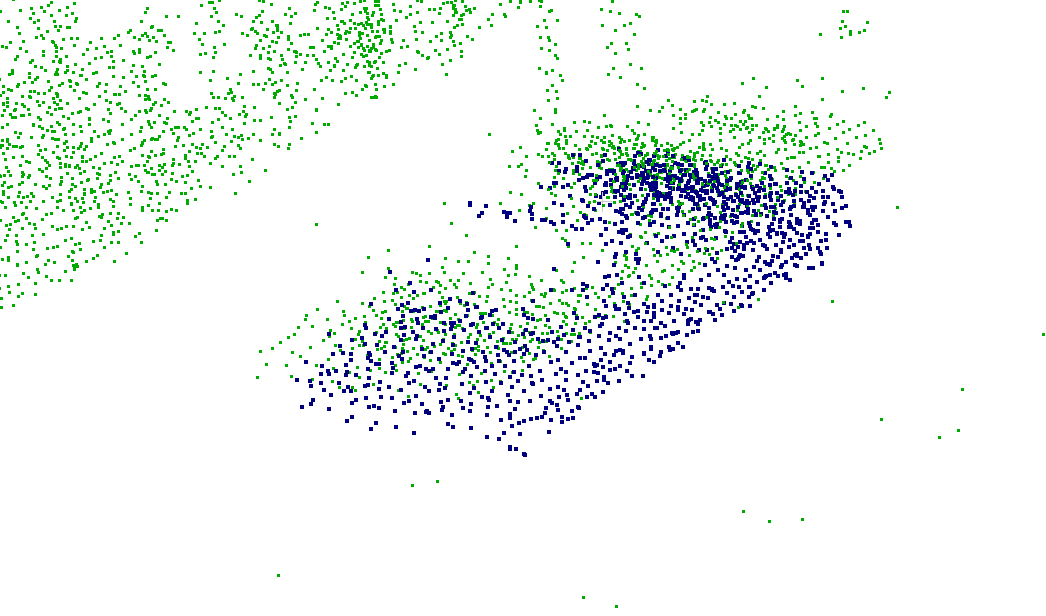} &
        \includegraphics*[width=0.44\linewidth]{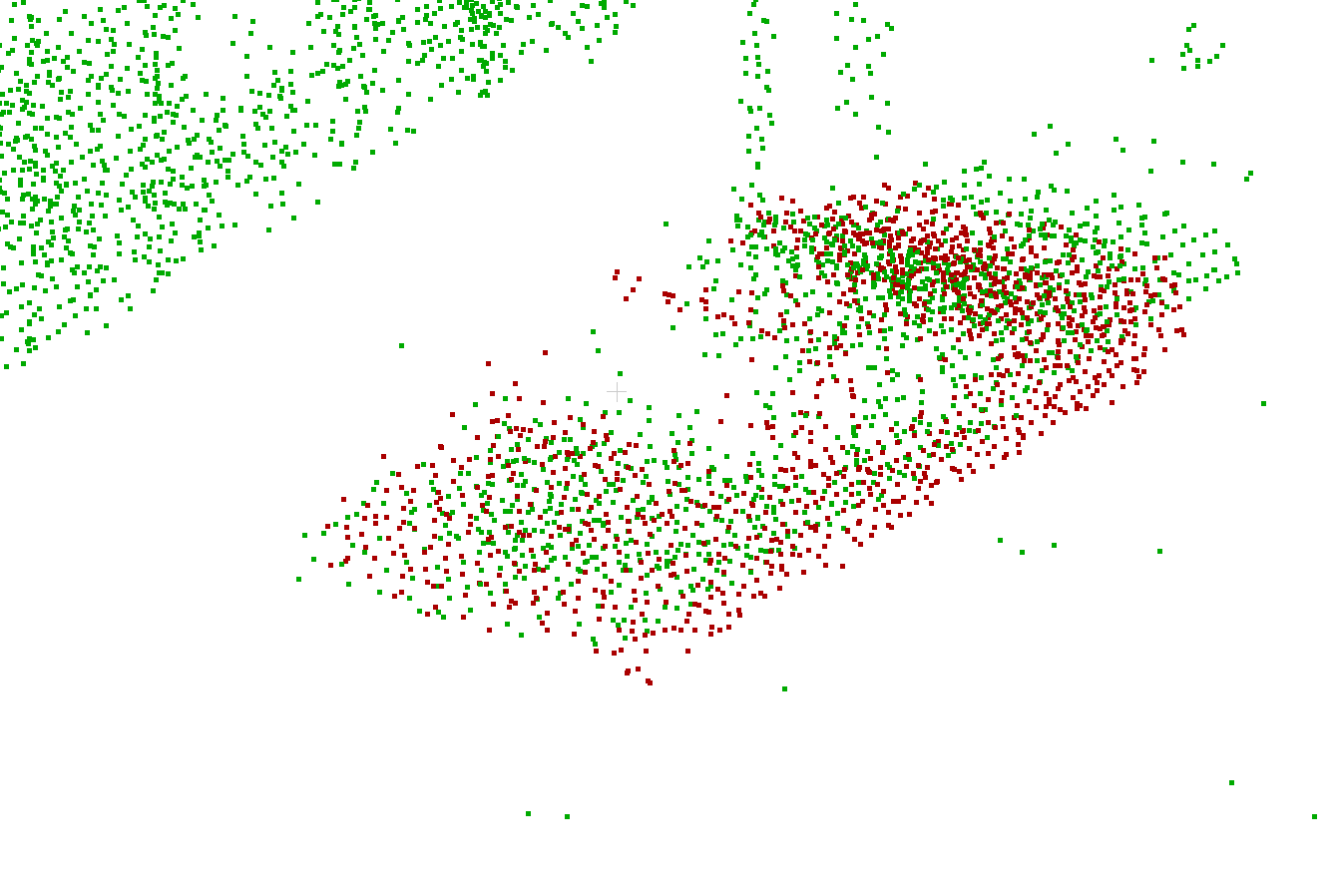} \\
         (c) & (d)\\
    \end{tabular}
        \caption{The required steps for aligning the local and target point cloud segments to cancel the drift in the odometry estimate: (a) we calculate the initial transformation by using the segment centroids; (b) we apply the RANSAC algorithm to filter the outlier matches; (c) we refine the update estimate by ICP matching between the corresponding segment point clouds. The result of the final alignment are shown in (d).}
    \label{fig:pose_refine}
    \end{center}
\end{figure}

For the pose alignment a straightforward method would be to use the segment centroids directly, but this might lead to errors c.f. Fig. \ref{fig:pose_refine}b-c. The reason for it is derived from the motion of the Lidar scanner. As the Lidar moves forward and detects a new segment for the first time, the form and line of sight of the object will be incomplete. Though the descriptors with eigenvalues or the CNN descriptor are robust enough to recognize an incomplete and a complete scan from the same objects \cite{segmap}, but the centroid of the incomplete segment will be shifted to the direction of the Lidar. In the update pose this could result in only a small transitional, but a huge angular error in the fitting, if we optimize for the minimum squared error. Instead, we used a more precise method using the entire point cloud of the segments. Let $W_e^W$ be the odometry pose estimate of the vehicle in the global reference frame \{W\} a dimension of 4x4 homogeneous transformation matrix and $W_t^W$ the true pose. Also $R_t^W$ is the 3x3 3D rotation matrix and $V_t^W$ is the 3x1 translation vector of $W_t^W$. Here $W_t^W$ can be obtained after calculating $W_u^W$ update transformation between the estimated and localized pose by

\begin{equation}
    W_t^W=W_u^W  W_e^W.
\end{equation}

Our only task is to calculate the update transformation following the steps below:
\begin{enumerate}
    \item We take the densified local point cloud $P_l^R$ of the robot as detailed in Section \ref{system_architecture}. \{R\} is a right-hand rule coordinate system in the \{W\}, where the position of the Lidar is in the origin and faces in $x$ axes direction. Here we first translate the cloud to the world coordinate system, to compare with the target cloud. In Fig. \ref{fig:pose_refine}(a) the local cloud is visualized by different colors for the different segments and the target map with white color. The vector connecting the matching segment pairs are drawn with a green line. On Fig. \ref{fig:pose_refine}(b) one can see the source cloud with orange and the target cloud with a green color now in the same coordinate system \{W\}. 
    \item For an accepted number of $n$ matches we calculate a prior mean transformation by transforming the $P_l^W$ by $W_p^W$ transformation matrix, where $R_p^W$ and the homogeneous row is an identity matrix, $V_{sp,i}^W$ is the vector connecting the centroids of the i\textsuperscript{th} segment pair and
    \begin{equation}
        V_p^W=\frac{1}{n}\sum_{i=0}^n V_{sp,i}^W
    \end{equation}
    The result of this prior transformation on a real segment can be seen in Fig. \ref{fig:pose_refine}(c) with the target cloud visualized with green color and a true positive matching segment by blue. 
    \item The final step of cloud alignment is an ICP step of point clouds. Here we only keep the points in $P_t^W$ target point cloud, that belongs to a true segment (accepted by the RANSAC filter) and align these set of points with an ICP step to the previously translated $P_l^W$. The output of this operation will now be a small refinement matrix $W_{ICP}^W$. The result of this fine alignment can be seen in Fig. \ref{fig:pose_refine}(d). 
\end{enumerate}

Finally $W_u^W$, and this way the truly localized pose of the robot can be calculated by simply following the multiplication order of our calculated transformation matrices, all defined in the world coordinate frame and resulting,

\begin{equation}
    W_t=W_u  W_e =  W_p W_{ICP}^{-1} W_e.
\end{equation}

This method was tested to be straightforward, with terminal statements of the ICP step computationally effective method, yet resulted in an excellent performance even running on an average strength PC and without the loss of tracking on the longest sequences. 

\section{EXPERIMENTS} \label{experiments}

All experiments were performed on a system equipped with
an Intel i7-6700K processor, with 32 Gb of RAM and an Nvidia GeForce GTX 1080 GPU performing easily in real-time operation. 
The timings for mean values in miliseconds and for standard deviations in parenthesis were 372.7 (7.2) for segmentation, 0.40 (0.26) for description, 26.27 (14.27) for match recognition in the reduced candidate pool, 0.09 (0.7) for the additinal RANSAC geometric filtering, and finally 88.0 (37.2) for the ICP alignment. The tested sequences were based on the Kitti Vision benchmark dataset, but we used especially the Drive number 18., 27. and 28. sequences as they were the longest in residential and city areas. The ground truth maps were created with the \textit{laser\_slam} tool that was made available with the SegMap algorithm. 

\begin{figure}[!t]
\centerline{\includegraphics[width=0.5\textwidth]{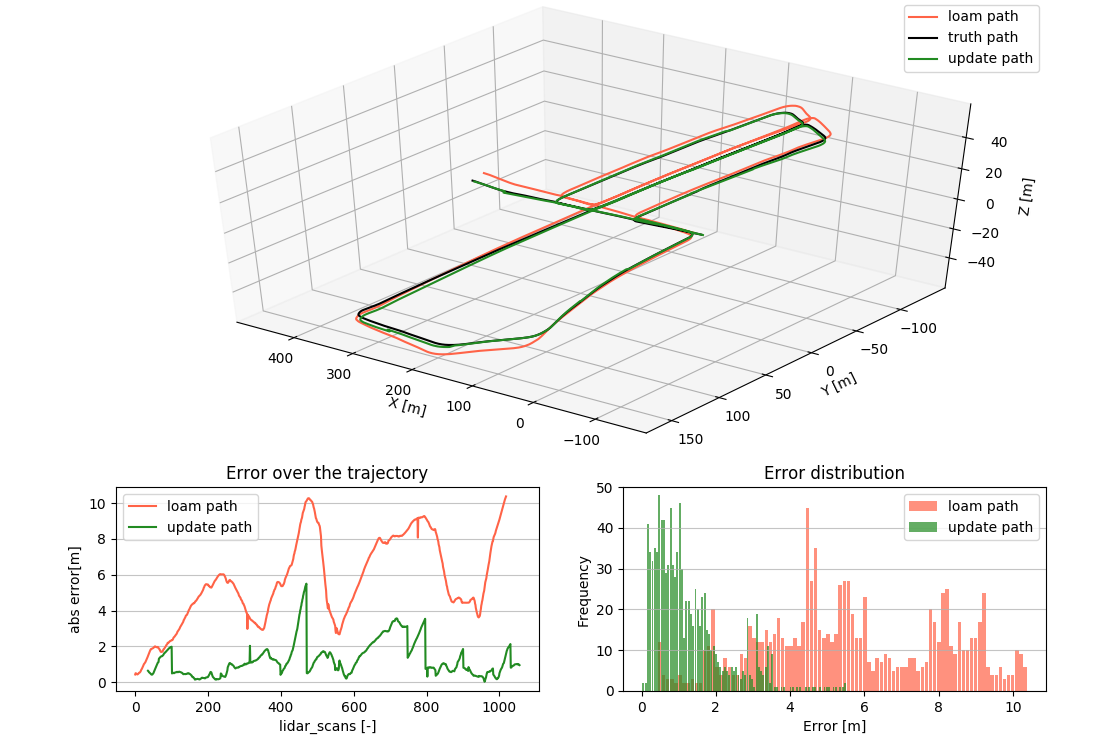}}\par
\caption{The absolute error for the relocalized poses of our updated path and the path from the original LOAM algorithm on the Kitti Drive 18 dataset. Here the value of the absolute error is calculated by comparing the global pose of a trajectory node at the timestamp of a scan to the corresponding ground truth trajectory pose with the same timestamp. The error is calculated by the 3D Euclidean distance of poses.}
\label{fig:abs_error}
\end{figure}

Fig. \ref{fig:abs_error} bottom diagrams are representing the quantitative results of our method compared to the original LOAM algorithm and the given ground truth trajectory. In Fig. \ref{fig:abs_error} left one can see the absolute error for the number of scans and their distribution in occurrence in Fig. \ref{fig:abs_error} right for the same sequence and color codes. Moreover, Fig. \ref{fig:abs_error} top shows the whole trajectory, where the black trajectory of ground truth is visualized together with the red odometry path and green path of our updated trajectory. As one can note from this or previously from Fig. \ref{fig:loam_maps} the resulted trajectory is significantly better than the original odometry without losing the ability of real-time performance and always being able to relocalize.

\begin{table}[!t]
\centering
\caption{Number of:  filtered out/true positive/false positive SegMap matches in case of different cluster sizes
}
\begin{tabular}{|l|c|c|c|c|}
\hline
Sequence\textbackslash Cluster size[-] & 2 & 3 & 4 & 5\\
\hline
Drive 18 & 579/49/2 &299/36/0 & 8/25/0 &0/22/0\\
Drive 27& 756/41/3 & 311/32/0 & 14/26/0 & 0/14/0\\
Drive 28 & 694/85/3& 421/70/0 & 158/54/1 &0/30/0\\
\hline
\end{tabular}
\label{tab:ransac_table}
\end{table}

In Table \ref{tab:ransac_table} we summarize the performance of our filtering method for the different cluster sizes of putative SegMap correspondence candidates and the examined three sequences. 
From the total number of putative SegMap correspondences the proposed LOL algorithm detected a number of: filtered out/true positive/false positive SegMap matches in case of different minimum cluster sizes. Here the filtered out matches were mostly containing true negative similarities. 
The validity of the matches were verified according to the correlation of their resulting transformation with the ground truth trajectory. Originally SegMap in localization mode did not included a similar filtering algorithm, only dealt with the problem of correspondence recognition. Based on Table \ref{tab:ransac_table} one can conclude that the proposed solution improved the relocalization significantly in a wide range of cluster sizes by filtering out false positive correspondence, thus incorrect localization updates. Also, one can establish the optimal cluster size as a hyperparameter of the localization problem. In our case one can see a cluster size of $2$ could still result in incidental false localizations, size of $5$ might result in too rare occurrences, but the size of $3$ or $4$ are a good balance between accuracy and frequency for robust localization. 

\section{CONCLUSION} \label{conclusion}
To conclude, this method solves the problem of Lidar-only odometry and localization in predefined 3D point cloud maps. Our algorithm consists of two state-of-the-art algorithms integrated in such a way to complement each other deficiencies and to highlight their advantages. Furthermore, we completed our solution with an advanced RANSAC filtering for additional certainty and ICP matching of the local- and target maps for increased precision.

\addtolength{\textheight}{-7.9cm}   




\bibliographystyle{IEEEtran}
\bibliography{lol}

\end{document}